\pdfoutput=1
\documentclass[conference]{IEEEtran}
\IEEEoverridecommandlockouts
\usepackage{cite}
\usepackage{amsmath,amssymb,amsfonts}
\usepackage{algorithmic}
\usepackage{graphicx}
\usepackage{textcomp}
\usepackage{xcolor}
\usepackage{url}
\usepackage{subcaption}
\usepackage{multirow}
\usepackage{booktabs}
\def\BibTeX{{\rm B\kern-.05em{\sc i\kern-.025em b}\kern-.08em
    T\kern-.1667em\lower.7ex\hbox{E}\kern-.125emX}}
\begin{document}
\title{DRPFNet: Dual-domain Residual Progressive Fusion Network for RGB-Thermal Object Detection}
\author{
Zian Wang, Changchun Li$^{*}$\thanks{$*$ Corresponding author.} \\
College of Computer Science and Technology, Jilin University, China \\
wangza2124@mails.jlu.edu.cn, changchunli93@gmail.com
}
\maketitle
\begin{abstract} 
RGB-thermal (RGB-T) object detection aims to fuse complementary information from visible and thermal modalities to achieve robust detection under varying illumination and weather conditions. Current methods typically employ attention mechanisms or transformers to perform cross-modal fusion independently at each feature scale, directly combining RGB and thermal features in the spatial domain. However, they still face significant limitations: 
cross-level knowledge inheritance caused by independent fusion at each scale, suppressing noise continuously due to the lack of bidirectional optimization, and information degradation induced by the absence of frequency-spatial collaboration. 
To address these issues, we propose DRPFNet, a Dual-domain Residual Progressive Fusion Network that constructs a unified information flow optimization system from three synergistic levels: structure, feature, and enhancement. At the structural level, we establish cross-scale propagation through bottom-up knowledge accumulation and bidirectional enhancement, ensuring smooth information flow. At the feature level, we collaboratively extract RGB high-frequency edges and thermal low-frequency structures via frequency band separation and edge guidance, guaranteeing representation quality. At the enhancement level, we enhance foreground-background discrimination through edge-guided dual-domain refinement, achieving precise object localization. Extensive experiments on two public RGB-T datasets demonstrate that our method achieves competitive performance with competitive efficiency, validating the effectiveness of this hierarchical collaborative strategy.
\end{abstract}


\section{Introduction}

RGB-thermal (RGB-T) object detection has emerged as a critical technology for all-weather autonomous systems, leveraging complementary information from visible and thermal modalities to achieve robust performance across varying illumination and weather conditions~\cite{ma2019infrared}. In recent years, it has witnessed substantial progress in multi-scale fusion and hierarchical feature interaction.

Despite these advances, current RGB-T detection methods still face several fundamental challenges that limit their effectiveness. A primary concern lies in the isolated nature of multi-scale fusion. Existing approaches~\cite{zhang2019cross,zhou2020mbnet,fang2022cft} typically perform cross-modal fusion operations independently at each feature level, treating different scales as separate entities rather than components of a unified hierarchical system. While these works establish solid foundations, this design philosophy inherently restricts the flow of information across scales---low-level geometric details cannot effectively propagate upward to enrich high-level semantic representations, and conversely, high-level global context remains unable to guide the selection and refinement of low-level features.

Building upon this, another critical limitation emerges from the predominantly unidirectional nature of current fusion-to-detection pipelines. Although various enhancement mechanisms have been proposed~\cite{li2019illumination,peng2023hafnet,liu2022target,wang2025ptmnet}, these methods generally follow a forward-only paradigm where fusion features flow directly into detection heads without mechanisms for iterative refinement. This unidirectional design constrains the system's ability to leverage detection-stage insights for improving fusion quality, ultimately limiting its capacity to continuously suppress modality-specific noise while preserving discriminative cross-modal information.

Additionally, most existing methods overlook the inherent spectral characteristics that distinguish RGB and thermal modalities. As established in prior work~\cite{ma2019infrared}, these two modalities exhibit natural complementarity in the frequency domain: RGB images predominantly capture high-frequency components containing fine-grained edge and texture details, while thermal images encode primarily low-frequency structural and contour information. However, prevalent fusion strategies~\cite{zhang2021gaff,shen2024icafusion,yang2025mmfn} operate exclusively in the spatial domain through attention mechanisms or Transformers, bypassing explicit frequency-domain processing. Recent empirical evidence suggests that such spatial-only approaches yield suboptimal feature representations~\cite{mcfusion2024} and struggle to effectively preserve the complementary frequency components inherent to each modality~\cite{zhang2025sgfnet}, resulting in unintended information degradation during cross-modal integration.

To address these interrelated challenges, we propose DRPFNet, a Dual-domain Residual Progressive Fusion Network that establishes a unified information flow optimization framework through three synergistic components. At the architectural level, we introduce the Multi-level Residual Fusion with Bidirectional Feature Enhancement (MRF-BFE) module to enable cross-scale knowledge propagation. This design implements progressive residual fusion where downsampled representations from preceding levels serve as geometric priors to guide semantic fusion at subsequent levels, while bidirectional enhancement pathways allow fused representations and modality-specific features to mutually refine each other through bidirectional feature refinement. At the feature extraction level, we develop the Dual-domain Adaptive Fusion (DFAF) module to jointly leverage frequency and spatial characteristics. This module employs learnable band separation to adaptively partition RGB high-frequency edges and thermal low-frequency structures in the frequency domain, while simultaneously applying Scharr operator-based edge guidance to preserve object contours in the spatial domain, with channel-wise gating dynamically balancing their respective contributions. At the detection enhancement level, we design the Edge-Guided Multi-scale Kernel (EGMK) module to enhance foreground-background discrimination. This component extends the dual-domain collaborative philosophy to the detection stage, utilizing edge-extracted attention maps to guide direction-sensitive convolutions toward object boundaries, while frequency-domain processing provides complementary structural information for effective foreground-background separation.

In a nutshell, the main contributions of this work are summarized as follows:

\begin{itemize}
    \item We propose DRPFNet, a dual-domain residual progressive fusion network, to addresses cross-scale propagation, bidirectional optimization, and frequency-spatial collaboration for RGB-T object detection.
    
    \item We design three key modules: Multi-level Residual Fusion with Bidirectional Feature Enhancement (MRF-BFE), Dual-domain Adaptive Fusion (DFAF), and Edge-Guided Multi-scale Kernel (EGMK).
    
    \item Extensive experiments on two public RGB-T datasets demonstrate that DRPFNet achieves competitive performance with competitive efficiency.
\end{itemize}

\section{Related Work}

\textbf{Cross-modal information interaction and fusion strategies.} The core challenge in RGB-T detection is effectively fusing complementary modality features. Early attention-based methods established the foundation~\cite{zhang2019cross,zhou2020mbnet,zhang2021gaff,li2019illumination,peng2023hafnet,liu2022target}, with CIAN~\cite{zhang2019cross} introducing cross-modality interactive attention and MBNet~\cite{zhou2020mbnet} addressing modality imbalance through differential fusion. Subsequently, channel-based fusion strategies~\cite{cao2023csaa,you2023msanet} emerged to dynamically select informative features from different modalities. Transformer adoption further enhanced global modeling capability~\cite{CDDFuse,shen2024icafusion,lee2024crossformer}, with CDDFuse~\cite{CDDFuse} pioneering dual-branch correlation-driven fusion and CrossFormer~\cite{lee2024crossformer} proposing cross-guided attention mechanisms. Recently, advanced fusion architectures demonstrate superior performance~\cite{yang2025mmfn,fusionmamba,wu2025wavemamba}, with MMFN~\cite{yang2025mmfn} proposing multidimensional fusion and Fusion-Mamba~\cite{fusionmamba} achieving linear complexity through state-space models. These methods demonstrate that cross-modal interaction capability continues to strengthen, yet still lacks cross-level knowledge propagation and bidirectional optimization mechanisms.

\textbf{Multi-domain feature representation and enhancement.} Frequency methods exploit inherent spectral differences between RGB and thermal~\cite{wu2025wavemamba,mcfusion2024,zhang2025sgfnet}. MCFusion~\cite{mcfusion2024} demonstrates that explicit frequency-domain processing yields superior features compared to spatial-only fusion, while SGFNet~\cite{zhang2025sgfnet} validates that frequency-domain separation effectively preserves complementary information. WaveMamba~\cite{wu2025wavemamba} employs wavelet transforms for multi-scale frequency decomposition, and subsequent works establish spatial-frequency learning frameworks via FFT-based approaches. For spatial alignment, existing works address geometric transformation and redundant spectrum removal~\cite{YUAN2024102246,zhao2024removal}. However, collaborative complementarity mechanisms between frequency and spatial domains remain insufficiently established.

\begin{figure*}[t]
\centerline{\includegraphics[width=\textwidth]{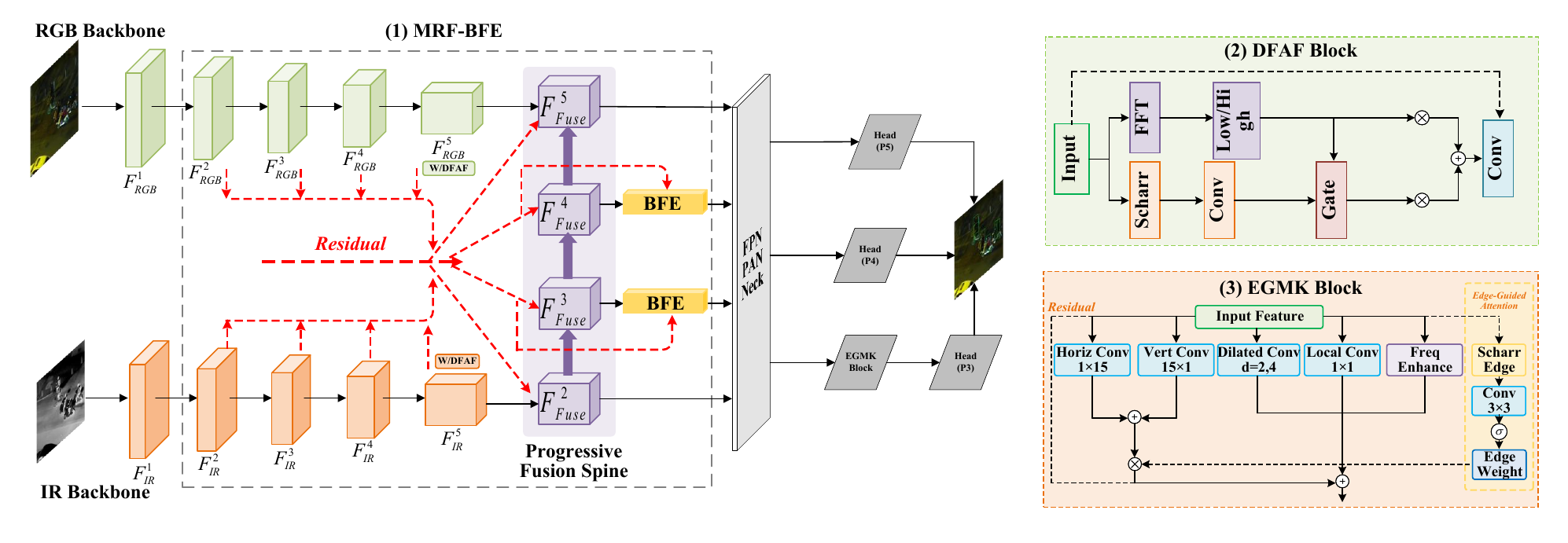}}
\caption{Overview of the proposed DRPFNet for RGB-thermal object detection. The network consists of three main components: (1) Dual-stream backbones with DFAF modules at each feature level for multi-scale feature extraction (simplified representation shown), (2) Progressive residual fusion with bidirectional enhancement (MRF-BFE) for cross-scale knowledge accumulation, and (3) Edge-guided multi-scale kernel (EGMK) module at P3 for foreground-background discrimination enhancement.}
\label{fig:architecture}
\vspace{-0.4cm}
\end{figure*}

\section{Methodology}

\subsection{Overview of DRPFNet}

DRPFNet adopts a dual-stream architecture consisting of three synergistic stages: feature extraction, progressive fusion with bidirectional enhancement, and detection, with our well-designed three key modules: Multi-level Residual Fusion with Bidirectional Feature Enhancement (MRF-BFE), Dual-domain Adaptive Fusion (DFAF), and Edge-Guided Multi-scale Kernel (EGMK), respectively. 

Specifically, in the feature extraction stage, two parallel YOLO11 backbones equipped with DFAF modules process RGB and thermal images separately, extracting modality-specific multi-scale features with both frequency and spatial domain information. In the progressive fusion stage, MRF performs bottom-up cross-modal fusion across four levels, where each stage concatenates dual-modality features with downsampled fusion results from the previous level for cross-scale knowledge accumulation. BFE then establishes reciprocal refinement at P3 and P4 levels by fusing original modality-specific features with corresponding fusion results. In the detection stage, the FPN-PAN pathway propagates features while incorporating bidirectional enhanced representations from both modalities, with the EGMK module deployed at P3 for foreground-background discrimination enhancement. 

The overall architecture is illustrated in Fig.~\ref{fig:architecture}.

\subsection{Dual-domain Adaptive Fusion Module (DFAF)}

The DFAF module integrates into the YOLO11 backbone to extract complementary information from both frequency and spatial domains (illustrated in Figure~\ref{fig:architecture}).

In the spatial domain, we employ the Scharr operator~\cite{scharr2000optimal} to extract edge features that capture object contours. We adopt the Scharr operator due to its superior rotational symmetry and reduced directional bias compared to the Sobel operator. The normalized edge features are processed through two convolutional layers to obtain spatial features $F_{\text{spatial}}$.

In the frequency domain, we apply 2D FFT to the input and separate the spectrum into low-frequency and high-frequency components through a learnable radius parameter. The low-frequency component contains global structural information, while the high-frequency component captures edge and texture details. Both frequency bands are transformed back to spatial domain via inverse FFT and fused through learnable weights:
\begin{equation}
F_{\text{freq}} = W[0] \odot \text{Low} + W[1] \odot \text{High}
\end{equation}

Finally, we fuse spatial and frequency features through a channel-wise gating mechanism. The concatenated features are processed by global average pooling and two $1 \times 1$ convolutions to generate gating weights $G$. The output is computed as:
\begin{equation}
Y = G \odot F_{\text{spatial}} + (1-G) \odot F_{\text{freq}} + X
\end{equation}
where the residual connection $X$ stabilizes training.

\subsection{Multi-level Residual Fusion and Bidirectional Feature Enhancement (MRF-BFE)}

\textbf{Progressive Residual Fusion.} Existing methods perform cross-modal fusion independently at each scale, lacking inter-scale information exchange. However, low-level features are rich in geometric details while high-level features contain semantic information---allowing low-level fusion results to participate in high-level fusion enables semantic understanding to be guided by geometric correspondences. Based on this insight, at each fusion stage $i \in \{2,3,4,5\}$ (corresponding to feature pyramid levels P2 through P5), we not only fuse the dual-modality features at the current level, but also introduce downsampled fusion results from the previous stage as prior knowledge:
\begin{equation}
\mathbf{F}^{i+1}_{\text{input}} = \text{Concat}(\mathbf{F}^{i+1}_{\text{RGB}}, \mathbf{F}^{i+1}_{\text{T}}, \text{DownSample}(\mathbf{F}^i_{\text{fused}}))
\end{equation}
The concatenated features are then enhanced through the C3k2 module~\cite{jocher2024yolo11}.

\textbf{Bidirectional Feature Enhancement.} In unidirectional fusion-to-detection pipelines, fused features cannot continuously leverage modality-specific discriminative information. To address this, we design bidirectional optimization pathways: in the forward path, fused features are concatenated with original modality features and enhanced, allowing fusion information to reinforce modality representations; in the backward path, the enhanced modality features are fed back to the FPN~\cite{lin2017feature} structure, achieving three-way fusion:
\begin{equation}
\mathbf{F}^3_{\text{head}} = \text{C3k2}(\text{Concat}(\text{Upsample}(\mathbf{F}^4_{\text{head}}), \mathbf{F}^{3,\text{enh}}_{\text{RGB}}, \mathbf{F}^{3,\text{enh}}_{\text{T}}))
\end{equation}

BFE is applied at P3 and P4 levels where modality-specific details are most discriminative, while P5 features are already highly semantic and benefit less from low-level enhancement. This design enables detection heads to simultaneously exploit high-level semantics, fusion information, and modality-specific details, forming a bidirectional refinement mechanism.

\subsection{Edge-Guided Multi-scale Kernel (EGMK)}

To extend the dual-domain collaborative philosophy from feature extraction to detection and enhance foreground-background discrimination, we deploy the EGMK module at the P3 detection level. While DFAF in the backbone focuses on modality-specific feature extraction, EGMK targets pre-detection refinement of the fused features.

In the spatial pathway, the Scharr operator extracts edge features to guide direction-sensitive strip convolutions~\cite{yuan2025strip} ($1 \times 15$, $15 \times 1$) toward object boundaries. In the frequency pathway, we employ learnable band separation unified with DFAF to extract low-frequency structural information. Cascaded dilated convolutions (dilation rates of 2 and 4) provide contextual information for distinguishing foreground from background regions.

EGMK is deployed specifically at P3 because this level contains the richest spatial details for boundary discrimination, while higher levels (P4, P5) are dominated by semantic features where edge-based refinement provides diminishing returns. The final features are aggregated as:
\begin{equation}
\mathbf{F}_{\text{out}} = (\mathbf{F}_{\text{h}} + \mathbf{F}_{\text{v}}) \odot \mathbf{A}_{\text{edge}} + \mathbf{F}_{\text{dilated}} + \mathbf{F}_{\text{local}} + \mathbf{F}_{\text{freq}}
\end{equation}
where $\mathbf{A}_{\text{edge}}$ denotes edge-guided attention maps, and $\mathbf{F}_{\text{h}}$, $\mathbf{F}_{\text{v}}$ represent horizontal and vertical convolution features.

\subsection{Complexity Analysis} 
Let the input feature map size be $H \times W$ with $C$ channels. The FFT/IFFT operations in DFAF have time complexity $O(CHW\log(HW))$ and space complexity $O(CHW)$; the remaining convolutions are $O(CHW)$. MRF-BFE only involves feature concatenation and $1 \times 1$ convolutions with complexity $O(CHW)$. EGMK employs depthwise separable convolutions, reducing the standard $k \times k$ convolution complexity from $O(Ck^2HW)$ to $O(C(k+C)HW)$; cascaded dilated convolutions achieve large receptive fields at $O(CHW)$. The overall complexity remains at the same order as the baseline detector.

\begin{table}[!t]
\centering
\caption{Performance Comparison on LLVIP Dataset. * denotes results from original papers.}
\label{tab:llvip_comparison}
\small
\setlength{\tabcolsep}{3.5pt}
\begin{tabular}{lc|cc}
\hline
\textbf{Method} & \textbf{Venue} & \textbf{mAP$_{50}$} & \textbf{mAP} \\
\hline
CFT~\cite{qingyun2022crossmodalityfusiontransformermultispectral} & arXiv'22 & \underline{97.5$^*$} & \underline{63.6$^*$} \\
MetaFusion~\cite{metafusion} & CVPR'23 & 91.0 & 56.9 \\
CSAA~\cite{cao2023csaa} & CVPRW'23 & 94.3$^*$ & 59.2$^*$ \\
Diff-IF~\cite{diffif} & IF'24 & 93.3$^*$ & 59.5$^*$ \\
ICAFusion~\cite{shen2024icafusion} & PR'24 & 95.2 & 60.1 \\
CAMF~\cite{camf} & TMM'24 & 89.0$^*$ & 55.6$^*$ \\
YOLO-Adaptor~\cite{yoloadaptor} & TIV'24 & 96.5$^*$ & --$^*$ \\
ACDF-YOLO~\cite{acdfyolo} & RS'24 & 96.5$^*$ & 61.3$^*$ \\
RsDet~\cite{zhao2024removal} & arXiv'24 & 95.8$^*$ & 61.3$^*$ \\
MDMDet~\cite{mdmdet} & ICHMS'24 & 96.5$^*$ & 61.2$^*$ \\
TFDet~\cite{xue2023tfdet} & TNNLS'25 & 96.0$^*$ & 59.4$^*$ \\
Fusion-Mamba~\cite{fusionmamba} & TMM'25 & 96.8$^*$ & 62.8$^*$ \\
MMFN~\cite{yang2025mmfn} & TCSVT'25 & 97.2$^*$ & --$^*$ \\
\hline
\textbf{DRPFNet (Ours)} & -- & \textbf{97.8} & \textbf{64.4} \\
\hline
\end{tabular}
\vspace{-0.2cm}
\end{table}

\section{Experiments}

\subsection{Experimental Setup}

\textbf{Datasets.} We evaluate our method on two RGB-thermal detection benchmarks: \textbf{M3FD}~\cite{liu2022target}: Contains 4,200 image pairs across 6 object categories. We follow the train/test splits from Liang et al.~\cite{Liang2023ExplicitAF}. \textbf{LLVIP}~\cite{jia2021llvip}: A challenging low-light visible-infrared paired dataset with 15,488 image pairs, containing a single category.

\textbf{Implementation Details.} DRPFNet is implemented in PyTorch and trained on a single NVIDIA RTX 3090 GPU with 24GB memory. We employ the SGD optimizer with an initial learning rate of 0.01, momentum of 0.937, and weight decay of 0.0005. The learning rate follows a cosine annealing schedule. The batch size is set to 4, with gradient accumulation applied. Input images are resized to 640$\times$640 pixels. Training epochs are set to 300 for M3FD and 100 for LLVIP. Standard data augmentation techniques are applied.

\textbf{Evaluation Metrics.} We adopt two standard metrics: mAP$_{50}$ is the mean Average Precision at IoU threshold of 0.5, and mAP is the mean Average Precision averaged over IoU thresholds from 0.5 to 0.95 with a step size of 0.05.

\subsection{Comparison with State-of-the-Art Methods}

\subsubsection{Results on LLVIP Dataset}
DRPFNet achieves 97.8\% mAP$_{50}$ and 64.4\% mAP on LLVIP, outperforming ICAFusion and other recent methods. This improvement stems from DFAF's ability to decouple modality-specific frequency characteristics: when visible light degrades in low-light scenarios, DFAF's learnable band separation preserves thermal low-frequency structures that encode object shapes independent of illumination. Simple spatial concatenation fails to exploit this spectral complementarity, causing information loss when RGB features become unreliable.

\subsubsection{Results on M3FD Dataset}
On M3FD, DRPFNet achieves 88.7\% mAP$_{50}$ and 61.8\% mAP, substantially outperforming ICAFusion~\cite{shen2024icafusion} and other recent methods. The improvement in mAP indicates better localization precision at higher IoU thresholds, which we attribute to MRF-BFE's progressive fusion: downsampled low-level features provide geometric priors that constrain high-level semantic fusion, reducing false positives from ambiguous detections. Bidirectional enhancement further refines object boundaries by allowing detection-stage gradients to back-propagate to modality-specific features, enabling continuous noise suppression.

\begin{table}[!t]
\centering
\caption{Performance Comparison on M3FD Dataset. * denotes results from original papers.}
\label{tab:m3fd_comparison}
\small
\setlength{\tabcolsep}{4pt}
\begin{tabular}{lc|cc}
\hline
\textbf{Method} & \textbf{Venue} & \textbf{mAP$_{50}$} & \textbf{mAP} \\
\hline
SuperFusion~\cite{superfusion} & JAS'22 & 83.5 & 56.0 \\
CFT~\cite{qingyun2022crossmodalityfusiontransformermultispectral} & arXiv'22 & \underline{88.2$^*$} & \underline{59.0$^*$} \\
TarDAL~\cite{liu2022target} & CVPR'22 & 80.5 & 54.1 \\
CDDFuse~\cite{CDDFuse} & CVPR'23 & 81.1$^*$ & 54.3$^*$ \\
IGNet~\cite{ignet} & MM'23 & 81.5$^*$ & 54.5$^*$ \\
ICAFusion~\cite{shen2024icafusion} & PR'24 & 87.1 & 57.0 \\
KCDNet~\cite{kcdnet} & TIM'24 & 83.2 & 56.3 \\
MRD-YOLO~\cite{mrdyolo} & Sensors'24 & 86.6$^*$ & 59.3$^*$ \\
CRSIOD~\cite{crsiod} & TGRS'24 & 84.0$^*$ & 57.2$^*$ \\
EMMA~\cite{emma} & CVPR'24 & 82.9$^*$ & 55.4$^*$ \\
RI-YOLO~\cite{riyolo} & YAC'24 & 83.6$^*$ & 56.6$^*$ \\
Fusion-Mamba~\cite{fusionmamba} & TMM'25 & 85.0$^*$ & 57.5$^*$ \\
MMFN~\cite{yang2025mmfn} & TCSVT'25 & 86.2$^*$ & --$^*$ \\
\hline
\textbf{DRPFNet (Ours)} & -- & \textbf{88.7} & \textbf{61.8} \\
\hline
\end{tabular}
\vspace{-0.2cm}
\end{table}

The advantage over Fusion-Mamba~\cite{fusionmamba} stems from architectural differences: progressive fusion explicitly propagates cross-scale features through residual connections, while state-space models process scales sequentially without geometric guidance. Similarly, the gains over transformer methods like CDDFuse and IGNet reveal that frequency-domain decomposition better preserves complementary information than spatial-only self-attention, which uniformly mixes modalities without exploiting spectral characteristics.
\begin{table}[!t]
\centering
\caption{Efficiency Comparison.}
\label{tab:efficiency}
\small
\setlength{\tabcolsep}{8pt}
\begin{tabular}{lcc}
\hline
\textbf{Method} & \textbf{Params (M)} & \textbf{Time (ms)} \\
\hline
ICAFusion~\cite{shen2024icafusion} & \textbf{120.2} & \textbf{26} \\
MMFN~\cite{yang2025mmfn} & 176.4 & -- \\
CFT~\cite{qingyun2022crossmodalityfusiontransformermultispectral} & 206.3 & 31 \\
Fusion-Mamba~\cite{fusionmamba} & 287.6 & 78 \\
\hline
\textbf{DRPFNet (Ours)} & \underline{134.9} & \underline{30} \\
\hline
\end{tabular}
\vspace{-0.2cm}
\end{table}

\begin{table*}[t]
\centering
\caption{Ablation Study on M3FD and LLVIP Datasets. MRF: Multi-level Residual Fusion, BFE: Bidirectional Feature Enhancement, DFAF: Dual-domain Adaptive Fusion, EGMK: Edge-Guided Multi-scale Kernel. Baseline employs dual-stream YOLO11 backbones with simple concatenation fusion at P3, P4, and P5 levels using independent C3k2 modules.}
\label{tab:ablation}
\small
\begin{tabular}{cccc|cc|cc|cc}
\toprule
\multirow{2}{*}{MRF} & \multirow{2}{*}{BFE} & \multirow{2}{*}{DFAF} & \multirow{2}{*}{EGMK} & \multicolumn{2}{c|}{M3FD} & \multicolumn{2}{c|}{LLVIP} & \multirow{2}{*}{Params} & \multirow{2}{*}{Time} \\
& & & & mAP$_{50}$ & mAP & mAP$_{50}$ & mAP & & \\
\midrule
$\times$ & $\times$ & $\times$ & $\times$ & 84.8 & 57.2 & 94.2 & 61.6 & 101.2M & 12ms \\
$\checkmark$ & $\times$ & $\times$ & $\times$ & 85.6 & 58.8 & 95.1 & 62.1 & 94.9M & 11ms \\
$\checkmark$ & $\checkmark$ & $\times$ & $\times$ & 86.7 & 60.3 & 95.7 & 62.9 & 99.9M & 12ms \\
$\checkmark$ & $\checkmark$ & $\checkmark$ & $\times$ & 87.8 & 61.1 & 96.7 & 63.8 & 123.0M & 26ms \\
$\checkmark$ & $\checkmark$ & $\checkmark$ & $\checkmark$ & \textbf{88.7} & \textbf{61.8} & \textbf{97.8} & \textbf{64.4} & \textbf{134.9M} & \textbf{30ms} \\
\bottomrule
\end{tabular}
\vspace{-0.3cm}
\end{table*}

\subsection{Efficiency Analysis}

Table~\ref{tab:efficiency} compares model efficiency. Compared to CFT, DRPFNet uses fewer parameters while maintaining comparable inference speed. Although requiring slightly more parameters and inference time than ICAFusion, DRPFNet achieves substantially better performance on both datasets as shown in Tables~\ref{tab:llvip_comparison} and~\ref{tab:m3fd_comparison}. Notably, it requires substantially fewer parameters than MMFN and Fusion-Mamba, demonstrating an effective balance between model capacity and detection performance.

\begin{table}[t]
\centering
\caption{Ablation Study on Bidirectional Enhancement Strategies on M3FD Dataset. Baseline configuration includes MRF, DFAF, and EGMK but without any bidirectional enhancement.}
\label{tab:bidirectional_ablation}
\small
\begin{tabular}{cc|cc}
\toprule
RGB Enhancement & Thermal Enhancement & mAP$_{50}$ & mAP \\
\midrule
$\times$ & $\times$ & 88.0 & 61.2 \\
$\checkmark$ & $\times$ & 88.2 & 61.2 \\
$\times$ & $\checkmark$ & 88.5 & 61.6 \\
$\checkmark$ & $\checkmark$ & \textbf{88.7} & \textbf{61.8} \\
\bottomrule
\end{tabular}
\vspace{-0.2cm}
\end{table}

\subsection{Ablation Studies}

\subsubsection{Component-level Ablation Analysis}
To validate the contribution of each proposed component, we conduct ablation experiments on both M3FD and LLVIP datasets by progressively removing modules from the full model. Table~\ref{tab:ablation} presents the results, where the baseline employs dual-stream YOLO11 backbones with simple concatenation fusion at P3, P4, and P5 levels. At each level, the concatenated RGB and thermal features are processed through independent C3k2 refinement modules. This baseline configuration lacks progressive residual connections across scales, bidirectional feature enhancement, dual-domain adaptive fusion, and edge-guided multi-scale kernels.

\textbf{Analysis.} The baseline has more parameters than the progressive configuration because it performs independent fusion at each scale with separate C3k2 modules, while progressive residual fusion employs lightweight residual connections. Starting from the baseline, progressive residual fusion with bidirectional enhancement brings substantial improvements in mAP, indicating better localization accuracy through geometric constraint propagation. Adding DFAF yields further gains via frequency-domain separation: RGB high-frequency edges guide localization while thermal low-frequency structures provide illumination-invariant shapes. EGMK contributes additional improvements through edge-guided refinement at P3, particularly effective for partially visible or occluded objects where edge precision is critical. 
\begin{figure}[t]
\centering
\begin{subfigure}[b]{\columnwidth}
    \includegraphics[width=\textwidth]{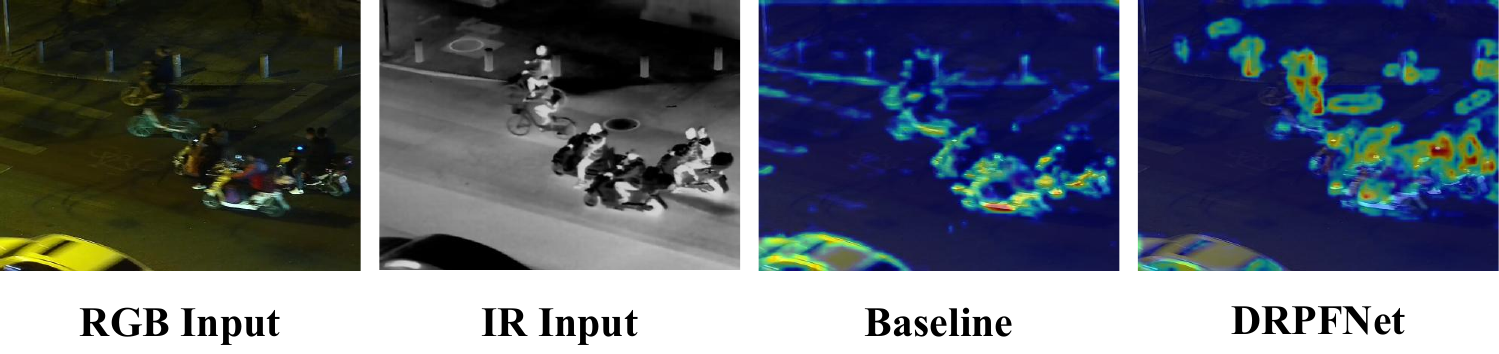}
    \caption{LLVIP dataset}
    \label{fig:heatmap_llvip}
\end{subfigure}

\vspace{1mm}

\begin{subfigure}[b]{\columnwidth}
    \includegraphics[width=\textwidth]{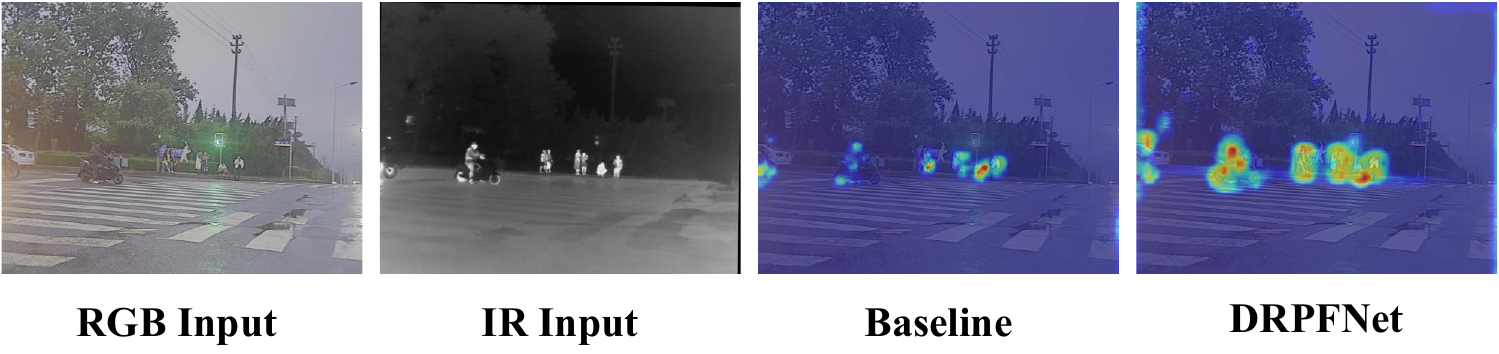}
    \caption{M3FD dataset}
    \label{fig:heatmap_m3fd}
\end{subfigure}
\caption{Grad-CAM visualization comparing baseline and full DRPFNet. On LLVIP, the baseline incorrectly focuses on electric scooters, while full DRPFNet shifts attention to pedestrian bodies. On M3FD, the baseline misses the partially visible car on the left, while full DRPFNet activates all targets.}
\label{fig:heatmap}
\vspace{-0.3cm}
\end{figure}
\subsubsection{Bidirectional Enhancement Ablation}
Table~\ref{tab:bidirectional_ablation} analyzes bidirectional enhancement strategies. Thermal-only enhancement outperforms RGB-only, confirming that thermal features remain reliable when visible light degrades. Full bidirectional enhancement achieves the best performance by enabling RGB features to refine object edges in well-lit regions while thermal features stabilize detections in dark areas.

\subsection{Visualization Analysis}

Figure~\ref{fig:heatmap} compares Grad-CAM activations between the baseline configuration and full DRPFNet, isolating DFAF and EGMK contributions.

On LLVIP, the baseline incorrectly activates electric scooters rather than riders. This occurs because spatial-only fusion cannot distinguish between thermally similar objects when RGB features degrade. DFAF resolves this by extracting thermal low-frequency structures that encode human body shapes distinct from vehicle geometries, enabling correct pedestrian localization.

On M3FD, the baseline misses occluded targets and weakly responds to distant objects. EGMK's edge-guided strip convolutions recover these targets by concentrating receptive fields along object boundaries, where edge features provide stronger discrimination than region-based features for partially visible objects.

\section{Conclusion}

In this work, we propose DRPFNet, a dual-domain residual progressive fusion network that addresses three fundamental limitations in RGB-thermal object detection: isolated multi-scale fusion, unidirectional optimization, and frequency-spatial collaboration deficiency. Through the synergistic design of MRF-BFE for cross-scale knowledge propagation, DFAF for dual-domain adaptive learning, and EGMK for foreground-background discrimination enhancement, our method achieves competitive performance on M3FD and LLVIP datasets while maintaining computational efficiency. However, performance may degrade when both modalities are simultaneously degraded, and evaluation on additional benchmarks remains future work.

\section*{Acknowledgment}
This work was supported by the College Students' Innovation and Entrepreneurship Training Program (Grant No. S202510183475).

\bibliographystyle{IEEEtran}
\bibliography{references}

\end{document}